\documentclass[conference]{IEEEtran}
\IEEEoverridecommandlockouts
\usepackage{cite}
\usepackage{amsmath,amssymb,amsfonts}
\usepackage{algorithmic}
\usepackage{graphicx}
\usepackage{textcomp}
\usepackage{hyperref}
\usepackage{xcolor}

\def\etal{\textit{et al. }}
\def\eg{\textit{e.g. }}
\def\ie{\textit{i.e. }}

\newcommand{\urlcolor}[1]{\textcolor{magenta}{#1}}

\def\BibTeX{{\rm B\kern-.05em{\sc i\kern-.025em b}\kern-.08em
    T\kern-.1667em\lower.7ex\hbox{E}\kern-.125emX}}
\begin{document}

\title{Automated Video Labelling: \\ Identifying Faces by Corroborative Evidence}

\author{\IEEEauthorblockN{Andrew Brown, Ernesto Coto, Andrew Zisserman}
\IEEEauthorblockA{\textit{Visual Geometry Group, Department of Engineering Science, University of Oxford} \\
\textit{Oxford, England}\\
\{abrown,ecoto,az\}@robots.ox.ac.uk}
}

\maketitle

\begin{abstract}
We present a method for automatically labelling all faces in video
archives, such as TV broadcasts, by combining multiple evidence
sources and multiple  modalities (visual and audio). We target the problem of ever-growing online video archives,
where an effective, {\em scalable} indexing solution cannot require a user to provide
manual annotation or supervision. To this end, 
we make three key contributions: (1) We provide a novel, simple,
method for determining if a person is famous or not  using image-search engines.
In turn this enables a face-identity model to be built reliably and robustly, 
and used for high precision automatic labelling; (2) We show
that even for \textit{less-famous people}, image-search engines
can then be used for \textit{corroborative evidence} to
accurately label faces that are named in the scene or the
speech; (3) Finally, we quantitatively demonstrate the benefits of our
approach on different video domains and test settings, such as
TV shows and news broadcasts. Our method works across three disparate
datasets without any explicit domain adaptation, and sets new
state-of-the-art results on all the public benchmarks.
\end{abstract}


\section{Introduction}
\label{sec:intro}
There has been an exponential growth in the volume of video content (in the form of TV and film material) being produced and released online. Such content is rich with useful information for researchers, historians and the general public. However, the sheer scale of the data, coupled with a lack of relevant metadata, makes indexing, analysing and navigating this content an increasingly difficult task. Relying on additional, manual human annotation is no longer feasible, and without an effective way to navigate these videos, this bank of knowledge is largely inaccessible. 

Interestingly, most video indexing and analysis is \textit{human-centric}. For example, we might want to navigate to a scene where two particular people interact, or where a group of people first appear together. This is partly because many videos in online archives are centred on humans, but also because of our natural interest in human actions and interactions. The focus of this paper is therefore the labelling of all faces in
videos, in a way that does not require any additional manual annotation from a user, be it in the form of provided transcripts~\cite{Everingham06a,Everingham09,8590759,Sivic09,bauml2013semi}, or a list of appearing people known \textit{a-priori}~\cite{Parkhi12b,Nagrani17b}. Our approach is hence scalable to large video archives where collecting manual annotations is infeasible.

\begin{figure}[t!]
\begin{center}
   \includegraphics[width=\linewidth]{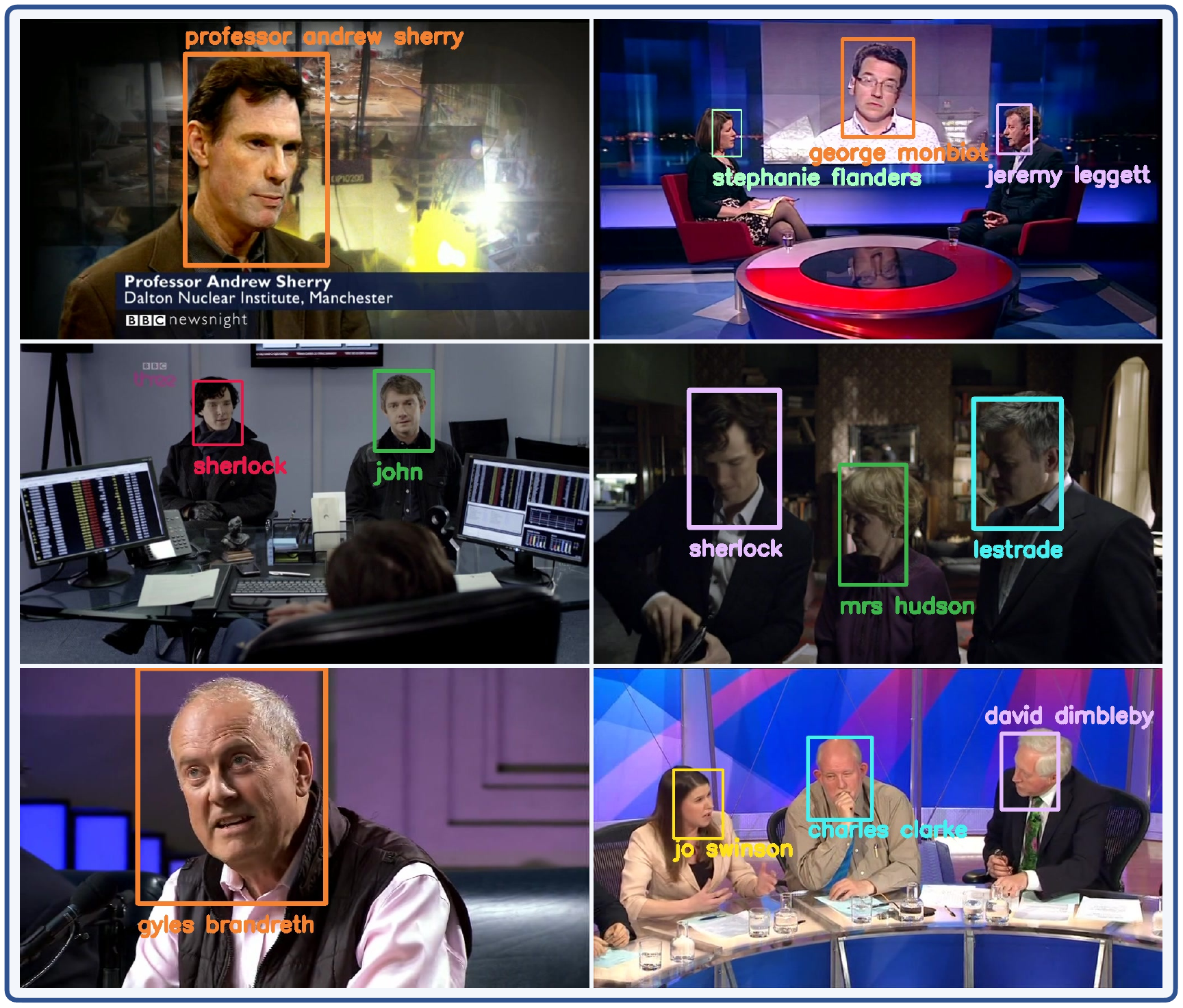}
\end{center}
\vspace{-2mm}
%\beforecaptions yeah it was bad 
\caption{
Modern, large, unlabelled online video archives, such as TV broadcasts, are growing at an exponential rate. These important resources are inaccessible due to the lack of annotations. Our proposed method for automated video labelling is scalable to these large archives, as it does not require additional manual supervision to label faces in videos. Examples from the BBC Videos and Sherlock dataset are shown above. The  method is able to label a wide range of people, over different domains, lighting conditions and in extreme poses.}
%\aftercaptions
\label{fig:teaser}
\vspace{-3mm}
\end{figure}

% To solve this task, we take a human-inspired approach. Imagine that you are watching a video and encounter a new person. \new{Perhaps that person is famous or not, but in order to confidently identify them with a name, you would rely on identifying clues in the video such as text on the screen, their name being mentioned in speech, or their presence in a list of cast members from an internet archive.} You might then find
% some evidence to \textit{corroborate} that this name is correct, by searching for the person online. 

To solve this task, we take a human-inspired approach. Imagine that you are watching a video and encounter a new person. In order to confidently identify them, you would first look for clues of their name either in the video such as text on the screen, their name being mentioned in speech, or in a list of cast members from an internet archive. You might then find
some evidence to verify that this name is correct, by searching for
the person online. In this work we follow this approach by harnessing the
freely-available weak annotations on the internet,
such as IMDB names lists and image-search engine rankings, to provide
evidence for recognising faces automatically and with confidence.
Consequently, this method is applicable to identities for whom
there are images online.

We denote
people with many images of themselves online as \textit{famous},
and introduce a novel approach for automatically identifying if an
identity is \textit{famous} without any additional manual
supervision or annotation. For identities that are
\textit{less-famous} according to our approach, we present a novel
method for using other identifying clues in the video together with
image-search engine results as \textit{corroborating evidence} to
provide confident person labels.

Beyond visual content, videos also contain identity information in the audio track, as humans can be recognised from the sound of their voice. Occasionally when used independently, neither face-appearance or voice can provide a confident identity label, be it due to a slightly obscured face, or a noisy or brief audio signal~\cite{Nagrani17b}. By fusing the two modalities and using a separate query expansion step, we show faces can be labelled with confidence, such that the recall of the automated labelling is improved without sacrificing precision. Figure~\ref{fig:teaser} gives example results of our automated labelling method. 

\begin{figure}[t!]
\begin{center}
   \includegraphics[width=\linewidth]{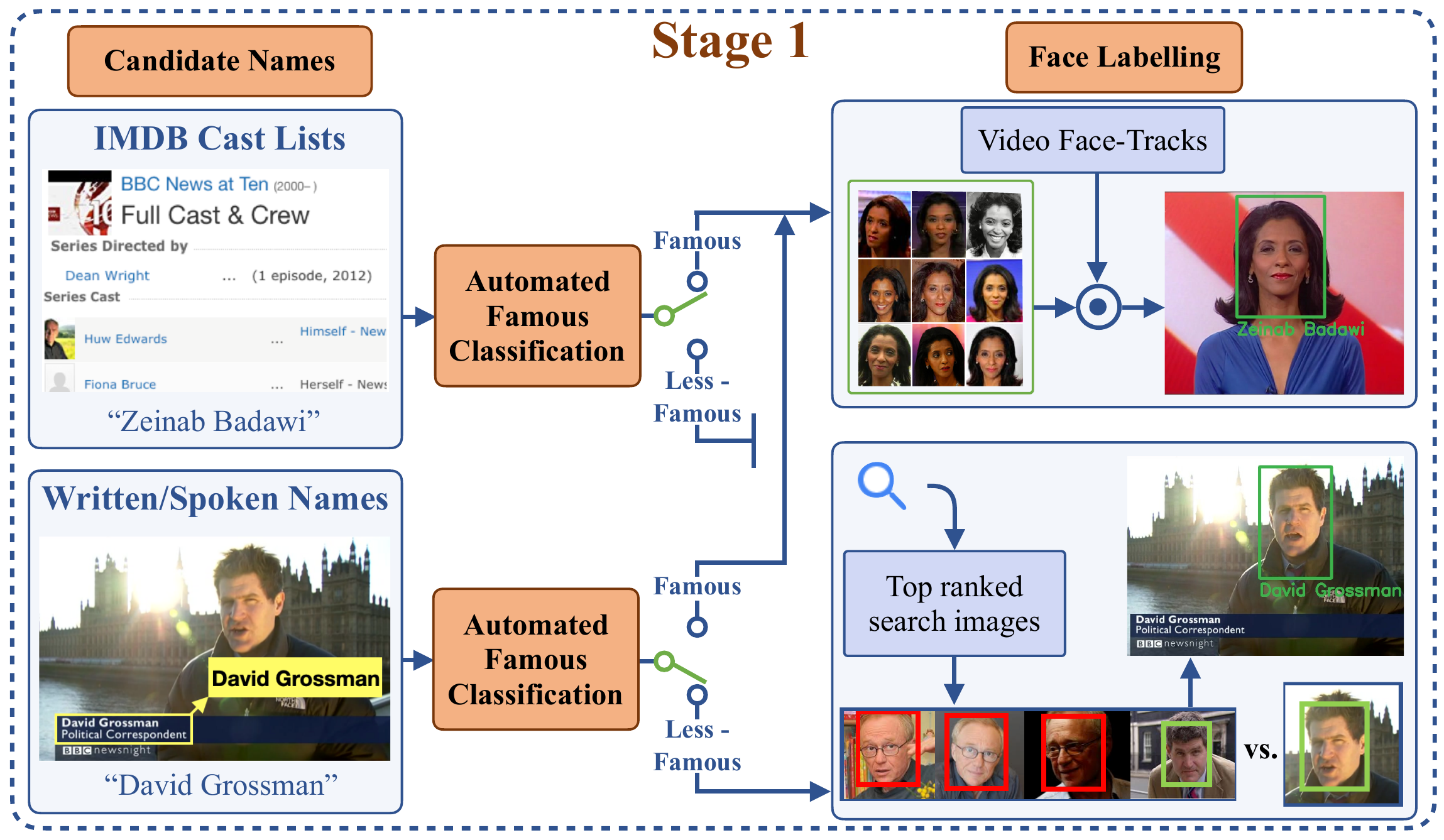}
\end{center}
\vspace{-3mm}
%\beforecaptions yeah it was bad 
% 
\caption{
The first stage of our method automatically finds candidate names and labels faces. Candidate names are
automatically sourced from IMDB name lists, displayed text
(\textit{written names}) and spoken words. Each person is classified as \textit{famous} or
not. For the \textit{famous} people, face-identity models are automatically
assembled and they are labelled throughout the videos. For the
\textit{less-famous} people who were found in written or spoken names, the temporal
occurrence combined with \textit{corroborating evidence} from image-search
engines is used to provide labels. }
%\aftercaptons
\vspace{-1mm}
\label{image_search_engine_figure}
\end{figure}

In summary, the task of this paper
is to assign identity labels (\textit{`tags'}) to all people in a set of test
videos, without the use of any additional manual annotation beyond what is freely and automatically available on the internet.
The base unit for labelling used in this paper is the face-track \ie
face-detections from consecutive frames of the same
identity that are linked together within shots.
% (\ie to indicate who is present in the videos) in order to make the
% process scalable to ever-growing online video archives. 
The  method consists of two key stages:

\noindent
\textbf{Stage 1 -- Using image-search engines as sole/corroborating evidence.}
This stage automatically identifies candidate names that may appear in the
test videos and classifies them as \textit{famous} or not. Image-search engines are used as  the sole source of evidence if they are \textit{famous}, or as \textit{corroborating evidence} if they are not in order to tag the faces. This is presented in
Section~\ref{sec:method} and shown in Figure~\ref{image_search_engine_figure}. 

\noindent
\textbf{Stage 2 -- Boosting the number of tags.}
This increases the number and variety of face tags across the test
videos, using two techniques: (i) fusing the information from the modalities of face-appearance and voice; and  (ii) query expansion. These techniques improve label
recall, while maintaining very high precision (see Section~\ref{tab_boosting}).

Our method for person labelling can proceed completely automatically starting from just the names of the programmes being labelled (and will obtain lists of appearing names in the process). We can also test specific modules of the method on standard benchmarks where cast lists are provided. 
We quantitatively demonstrate the benefits of our approach on several different video domains and test settings, such as TV shows and news broadcasts, as described in Section~\ref{sec:experiments}. The results in Section~\ref{sec:results} show that our method works across these disparate datasets without any explicit domain adaptation, and sets new state-of-the-art results on all the public benchmarks. Further details can be found at \urlcolor{\url{https://www.robots.ox.ac.uk/~vgg/research/person_id_in_video/}}.

\section{Related Work}
\label{sec:related_work}
\noindent
\textbf{Labelling People in Videos:}
This task has been well studied
\cite{Nagrani17b,Parkhi12b,Everingham06a,Everingham09,8590759,bauml2013semi,haurilet2016naming},
due to its uses for story understanding~\cite{Bain20} and archive indexing. In
previous works, different levels of prior information are assumed to
be available: Everingham \etal \cite{Everingham06a} make use of
transcripts aligned with subtitles to provide weak supervision, with
many other works following suit
\cite{Bojanowski13,Cinbis11,Cour10,Parkhi15a,Sivic09,Tapaswi12,Everingham09,8590759,bauml2013semi}. 
Often the task is posed as one of Multiple Instance Learning
(MIL) \cite{haurilet2016naming,8590759,Bojanowski13,Kostinger11,Wohlhart11}.
Nagrani
and Zisserman \cite{Nagrani17b} instead presume the existence of
cleaned web-downloaded images for actor-level supervision. 
Many of these methods are tasked with labelling only a small number of known,
main characters, and either require transcripts or some additional manual
annotation in the pipeline. Our method, on the other hand, does not
require either, and so crosses the domain gap to
real-world large video archive scenarios. 

The labelling of people in news videos is also a well studied and challenging task, due to its
open-ended nature, where a list of appearing people or transcripts is
not
available~\cite{satoh1997name,le2017towards,yang2005multiple}. Both
Canseco \etal \cite{canseco2005comparative} and Mauclair \etal
\cite{jousse2009automatic,mauclair2006speaker} establish
correspondences between spoken names and speech-turns to label
people. Several works use the co-occurrence of overlayed names in a
scene with speech-turns to make person labels
\cite{poignant2014unsupervised,gay2014comparison,bendris2013unsupervised,8880535}, with much of this area of research
accelerated by the MediaEval ``Person Discovery in Broadcast TV''
challenges \cite{poignant2017multimodal}. However, many of these
techniques rely upon heuristics based on TV-broadcast structure to
make confident labels. These heuristics do not generalise well to other
domains. Our method on the other hand does not rely on such
heuristics, and can hence be applied to movies, TV material and
broadcast news alike.

\noindent
\textbf{Using Image-Search Engines for Supervision:} Although widely used in the Vision Community for object recognition \cite{Chatfield12,Fergus05a,Schroff07,Berg06,lin2003web}, and more recently for face recognition \cite{Nagrani17b,Parkhi12b,Parkhi15,le2008unsupervised,holub2008unsupervised,pham2010naming,chen2019name,Brown20}, the problem faced is that retrieved results have varying precision. Previous person-identification work has either relied on manually removing false positives from retrieved images \cite{Nagrani17b}, or has focused on automatically improving the precision \cite{Parkhi15,le2008unsupervised,holub2008unsupervised}, starting with a pre-determined list of well-known people. Both are infeasible in real-world video archive scenarios. In this work, to our knowledge, we present the first method for automatically determining the usefulness of a set of search-results, before automatically removing outliers from the retrieved results of those deemed useful, resulting in a completely automatic, scalable and high-precision method.

% \noindent
% \textbf{Fusing Evidence:} \todo{this needs to be better} Previous work on fusing evidence from voice and face signals for person recognition can be broken down into feature-level fusion \cite{liu2018iqiyi,shon2019noise,mclaughlin2013robust} and late-fusion \cite{ekenel2007multi,choudhury1999multimodal,luque2006audio,4959999,sell2018audio,Nagrani18c} techniques. Liu \etal \cite{liu2018iqiyi} propose attention based techniques on the feature level to take into account the importance of each modality, where as Choudhury \etal \cite{choudhury1999multimodal} simply propose a weighted combination of the scores from each uni-modal network. Most similar to our work are those that try to explicitly take into account the quality of either modality during fusion \cite{shon2019noise,mclaughlin2013robust, furnari2019would}. 

\section{Stage 1 -- Using image-search engines as sole/corroborating evidence}
\label{sec:method}

In this section, we describe our approach for using image-search engines to obtain sole or  \textit{corroborating evidence} for labelling people in videos. As a precursor, there are three cases for the usefulness of image-search engines depending on the candidate name: if the person is \textit{famous}, many images of them will be found online (retrieved results will have high precision); for the \textit{less-famous} people, there may be one or two images of them returned;  and for the \textit{never-famous} people there will be none (retrieved results have zero recall). 

This stage of the method proceeds in two steps. The first is {\bf obtaining candidate names.}
The names of appearing people are not known \textit{a-priori} and
therefore must be first automatically gathered before any labelling
can commence. For video material like TV news broadcasts or movies, we use three different sources: (i)
IMDB names lists (ii) Scene text: names appearing in text
displayed on screen during the programmes, such as overlaid banners. (iii) Speech:
names appearing in the speech of the audio track.
This step results in a set of candidate names for each episode.
Details of the fully automatic processing to obtain these candidate names are given in
Section~\ref{sec:vid_pre_processing} and the Supplementary Material.

\begin{figure}[t!]
\begin{center}
   \includegraphics[width=\linewidth]{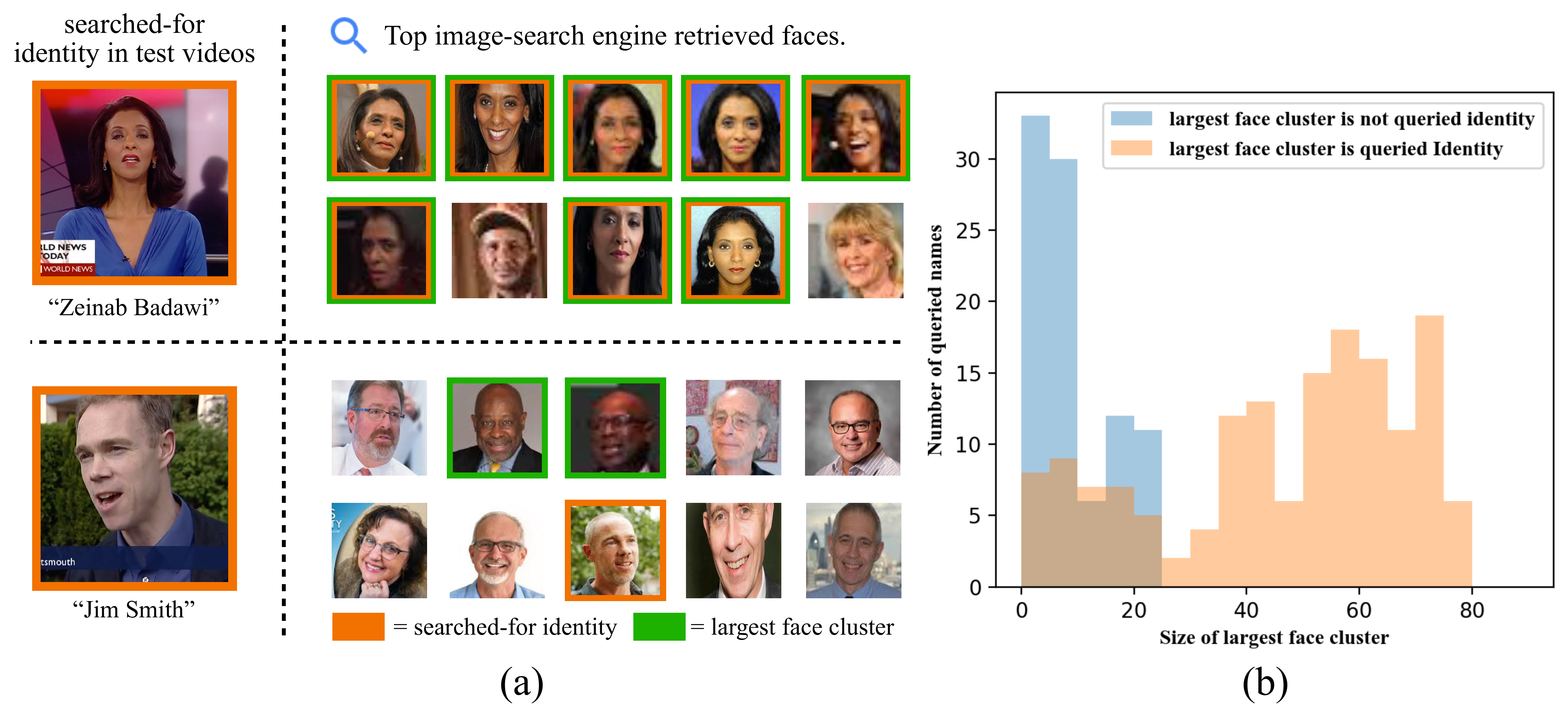}
\end{center}
\vspace{-4mm}
%\beforecaptions yeah it was bad 
\caption{(a) Examples of the top-10 retrieved image-search engine results for two different candidate names. For the ``Zeinab Badawi'' query (top row), the large face cluster (shown in green) has 8 faces, and all depict the queried person (shown in orange). The actual cluster for ``Zeinab Badawi'' in the top-100 retrieved images has 76 faces.  For the ``Jim Smith'' query (bottom row), the largest face cluster is far smaller with 2 images, and does not depict the searched-for person. The observation is made that if the largest face cluster has many images, then it will likely contain the searched-for person. (b) A plot of whether the largest face cluster contains the searched-for identity for 350 randomly chosen candidate names found for the BBC Videos dataset. Clearly, if the largest face cluster has many faces in it (\eg more than 30 in the top-100) then it is very likely to contain the searched-for person. }
% 
%\aftercaptions
\label{search-engine-cluster}
\end{figure}

The second step is {\bf  labelling \textit{famous} and \textit{less-famous} people.}
Given a candidate name, we first determine if the person is \textit{famous} or not 
using a novel method explained below based on downloaded images from an image-search engine. 
We then proceed in one of two ways:
(1) For each \textit{famous} name, a face model is built from the downloaded images and used as the sole evidence to label that person throughout the video; or  (2)
For the \textit{less-famous} people, we use the temporal occurrence of
their spoken or written (displayed) name in a video as primary evidence, and any
single occurrence of them in retrieved image-search engine results as
\textit{corroborating evidence}.
There is no labelling method for the \textit{never-famous} people, as image-search engines provide no examples of their appearance.
The following describes  these methods in detail.
% Once again all methods used require no manual annotation or supervision, or lists of appearing people known a-priori. 

% 
\noindent
\textbf{Sourcing Candidate Names from IMDB:} 
The name lists are freely and automatically
obtained starting from just the name of the programme (which often can
be found automatically in video metadata). These lists do not constitute curated
cast lists as they often contain
thousands of names of briefly appearing characters. This differs from
previous non-scalable work on automated face labelling, e.g.~\cite{Nagrani17b,Parkhi15,Parkhi12b},
which use a curated list of \textit{famous} appearing names.

\noindent
\textbf{Classifying Candidate Names as \textit{famous}:} 
When a candidate name is queried in the image-search engine, we use the key observation that if many of the top-ranked retrieved results correspond to the same person, then this person is \textit{famous}. This observation is illustrated in Figure~\ref{search-engine-cluster}.
% person is the queried person, and the 
% This
% hypothesis can be simply verified with a simple experiment showing
% whether the largest cluster of identities corresponds to the
% searched-for person, and the size of the cluster, as shown in figure
% x(to do). As can be seen, above a certain cluster size, the depicted
% identity almost always corresponds to the queried person, with the
% only false positives corresponding to when several \textit{famous}
% people have the same name. 
In detail, faces are detected in the top 100 ranked results, and clustered using 
Agglomerative-Clustering \cite{jain1988algorithms} on their
L2-normalised face embeddings~\cite{Cao18} (using a cosine-distance threshold of 0.7).
If
the largest face-image cluster has more than 
$\alpha$ faces (in this work we use $\alpha = 30$, learnt on a validation set as described in Section~\ref{famous_non_famous_results}, 
and also illustrated in Figure~\ref{search-engine-cluster}),  then the identity is classified as \textit{famous}.
\noindent
\textbf{Building a Face-Identity Model For the \textit{famous} People:}
For the candidate names that are classified as \textit{famous}, we simply
build a biometric model for that identity and use it to label any
face-tracks depicting that person in all test videos, as shown in
Figure~\ref{image_search_engine_figure}. We take the largest cluster
of face-embeddings from the downloaded images and average-pool and L2-normalise them into a single embedding. This single embedding is now a face-identity model that can be used for labelling. Taking only the embeddings from the
largest cluster serves the purpose of removing false positives from
the downloaded images. Face-tracks in the test videos are then labelled by measuring the cosine similarity between their embeddings and the face-identity embedding. If a face-track embedding has a similarity score higher than a threshold learnt on a validation set, then it is labelled with the famous name.

% No classifiers are trained
% in this work for scalability purposes, so that further candidate names
% can be continuously added without the need for any further training.

% 
\noindent
\textbf{Finding Corroborating Evidence for \textit{less-famous} People:} 
When a candidate name is not classified as \textit{famous}, there may still be a few images of the person in the downloaded images (\eg the bottom example in Figure~\ref{search-engine-cluster}). These low-precision images cannot serve as the sole evidence as was the case for the \textit{famous} people, but can serve as \textit{corroborating evidence}. Hence, for \textit{less-famous} people we use the
temporal occurrence of their spoken or written name as primary
evidence of their appearance, and then a single correct retrieved
face from image-search engines as the necessary \textit{corroborating
evidence} to label.  In detail, the \textit{corroborating evidence} is that at
least one of the 20 top ranked faces from an image-search engine matches the face-track that appeared in the test video when the name was found. The face-track is then labelled with that name. We are here using 1-to-1
face verification. This is
less accurate than the template-based (many-to-one) face verification
that we use for the \textit{famous} names, however seeing as the evidence is supported by
the presence of the name in the scene or audio track, it is sufficient here. 
\section{Stage 2 --  Boosting the Number of Tags} \label{tab_boosting}
In this section, we describe the two methods used for boosting the number and variety of tags in the test videos. This includes fusing the evidence sources of face-appearance and speech, as well as query expansion.

\subsection{Fusing face-appearance and voice} \label{fusion}
% 
% In this section, we explain the query expansion step in the proposed method. 
This section explains how we use additional information from the speech modality as corroborative evidence to label further face-tracks when the face-appearance alone is not enough to make a confident tag. For each tagged face-track, we use Active Speaker Detection~\cite{Chung16a,8682524} (ASD), to classify whether the face is speaking. For the speaking faces of each tagged person, we extract temporally aggregated speaker embeddings~\cite{chung2020in} using the overlapping audio segments, which after average pooling form a speaker model for that identity. For the remaining un-labelled speaking face tracks in the video, we compute the similarity score between the speaker embedding and the speaker ID models (voice score), and the similarity score between face and the face ID models (face score). We then simply average the two scores (fusion score), and label the face-track if it is above a given threshold.   We find empirically that the simple rule of averaging the two scores is highly effective. For any speaking face-track to be incorrectly classified with the fusion score, both the voice and the face score needs to be high for the same, incorrect identity. We see during experiments that this is very rarely the case, due to the lack of coupling between the modalities. In the supplementary material we present experiments with more complex architectures and show that this very simple rule achieves comparable performance. In the next sections we refer to this method as ``Stage 2 Fusion".
\subsection{Query Expansion} \label{query_expansion}
% 
% In this section, we explain the query expansion step in the proposed method. 
Query expansion (QE) is a popular re-ranking method \cite{Chum07b,Arandjelovic12,Buckley95,Salton99}. Methods assume top ranking instances to be from the same class as the query, and use these to supplement the original query to create a new, superior ranking. Nagrani and Zisserman~\cite{Nagrani17b} perform QE by training a new classifier for each identity with their top ranked test-video tags, and show it to be helpful for crossing the domain gap between the search engine face-images and the TV-material face-tracks (similarly explored in~\cite{Brown20b} for voice). We adopt the same technique in this work, except we do not train any additional parameters at this stage, but instead just average-pool all tagged face-image embeddings to form a new face-identity model, which is then used to make further tags. In the next sections we refer to this method as ``Stage 2 QE".

\section{Datasets, Evaluations and Implementation}
\label{sec:experiments}
In this section we first describe the datasets and the evaluations used for assessing the method, and then give some details on implementation. The video labelling method can proceed completely automatically given only the programme names. This `plug and play' automation is assessed in experiments on two datasets: BBC Videos and MediaEval. We also test out Stage 2 of the method on the standard person identification benchmark, Sherlock,  where cast lists and corresponding face-images are provided. Statistics of the three datasets used are given in Table~\ref{tab:dataset_table}. Different test protocols are used, when either testing the whole method, or just Stage 2, so that previous published methods can be compared to.

\noindent
\textbf{BBC Videos:} This dataset consists of five episodes of different BBC television programmes (BBC
news, BBC World News, Newsnight, Question Time). The challenging dataset for identification provides annotations for all \textit{human-identifiable}  characters, from road-side interviewees to well-known politicians.  These are the people
whose names are alluded to somewhere in the episode, or
who are well-known \ie which a human-annotator with access to the internet could annotate. This includes most people barring audience members, pedestrians, etc. Only the names of the programmes are provided to the method. The first episode constitutes the validation set.

\begin{table}[t]
\caption{Dataset Statistics and information on which parts of the method are tested on each of the datasets.  Annotations are either provided at the face-track level (BBC Videos, Sherlock) or at the scene level (MediaEval).}
\label{tab:dataset_table}
\centering
\begin{tabular}{|l|c|c|c|}
\hline
\textbf{Dataset} &\textbf{BBC Videos}& \textbf{MediaEval~\cite{poignant2017multimodal}}& \textbf{Sherlock~\cite{Nagrani17b}}  \\ \hline
No. Identities          & 66                    & 1,971                                                                                   & 31                                                                                                                                                \\
No. Annotations                & 1,971                 & 6,889                                                                                   & 5,246                                                                                                                                             \\
No. Videos                        & 5                     & 79                                                                                      & 3                                                                                                                                                \\
No. Hours                         & 2.2                   & 49.1                                                                                    & 4                                                                                                                                                 \\ \hline
Test Stage 1                     & \checkmark & \checkmark      &                                    \\
Test Stage 2 - Fusion              & \checkmark & \checkmark    & \checkmark                                        \\
Test Stage 2 - QE                  & \checkmark & \checkmark    & \checkmark  \\ 

\hline
\end{tabular}
\vspace{-3mm}
\end{table}

\noindent
\textbf{MediaEval~\cite{poignant2017multimodal}:} This dataset featured in the MediaEval 2015 challenge  ``Multimodal person discovery in broadcast TV" \cite{poignant2017multimodal}. The test set \cite{mediaevalgithub} consists of 79 episodes from the French TV news show, \textit{“Le 20 heures”}, with a total of 6,889 speaking faces annotated at the scene level. Only the name of the programme is provided to the method. 

% The face-track level labels that are provided by the proposed method are simply extrapolated to scene level labels for evaluation.}

% This dataset featured in the MediaEval 2015 challenge: Multimodal person discovery in broadcast TV \cite{poignant2017multimodal}. Over a set of provided textual queries (supplied as a list of strings), the task is to retrieve the shots where that person is both speaking and is visible,  with no list of appearing people provided before-hand. Performance is measured by mean Average Precision. The development set is composed of 137 hours of various TV shows,  and the publicly available version of the test set \cite{mediaevalgithub} consists of 79 episodes from the French TV news show, \textit{“Le 20 heures”}. The challenging test set asks for 1,971 unique queries to be found across 6,889 total annotations, ranging from frequently appearing famous anchors, to far less famous road-side interviewees.

\noindent 
\textbf{Sherlock~\cite{Nagrani17b}:} This dataset consists of three episodes of the crime drama show ``Sherlock", where the face-tracks depicting main characters have been annotated. Each episode is 
approximately  80 minute long. A cast list and face-identity models (in the form of image-search engine images) are provided for each of the annotated characters, and the task is to classify each face-track by identity. The fast-paced show contains many visually challenging scenes (\eg dark, quick camera movement), making it a difficult task for person labelling.
For fair comparison to previous works, the face-tracks provided with the Sherlock dataset are used in the experiments.

% The fast-paced show contains scenes of vastly different lighting and viewpoints, making it a challenging task for identification

% with the task being to maximise the classification accuracy of a large number of unique cast members annotated throughout. The fast-paced show contains scenes of vastly different lighting and viewpoints, making it a challenging task for identification, and has been previously used by the community~\cite{Nagrani17b} to test the use of web-search images for labelling people in videos. For fair comparison to the previous state-of-the-art, we use the same list of names, downloaded web-search images and test protocol that was employed by Nagrani \& Zisserman~\cite{Nagrani17b}, and hence use the experiment to text stage 2 of the method. These web-search images were manually cleaned by the original authors. 

% \noindent
% \textbf{IQIYI-Fusion:} This dataset is a subset of the IQIYI-VID-2019 Celebrity Video Identification dataset~\cite{IQIYI_ACM}, containing only videos of speaking faces for fair evaluation of a module fusing face-appearance and voice. Each clip in the dataset is an average of 3.9 seconds long. The task is to correctly classify each of the test clips as one of the identities from the training clips. Full details of the motivation and construction behind IQIYI-Fusion are contained in the supplementary material.

\noindent
\textbf{Evaluation Metrics:} When starting from the programme name alone, the BBC Videos and MediaEval experiments constitute open-set retrieval tasks, and so are evaluated using retrieval metrics: Precision, Recall, mAP, class recall  -- a measure of how many of the total classes have had 1 instance correctly retrieved. The dataset testing Stage 2 alone (Sherlock) provides a closed-set classification task, and so is evaluated using classification accuracy.

\begin{figure}[t!]
\begin{center}
   \includegraphics[width=\linewidth]{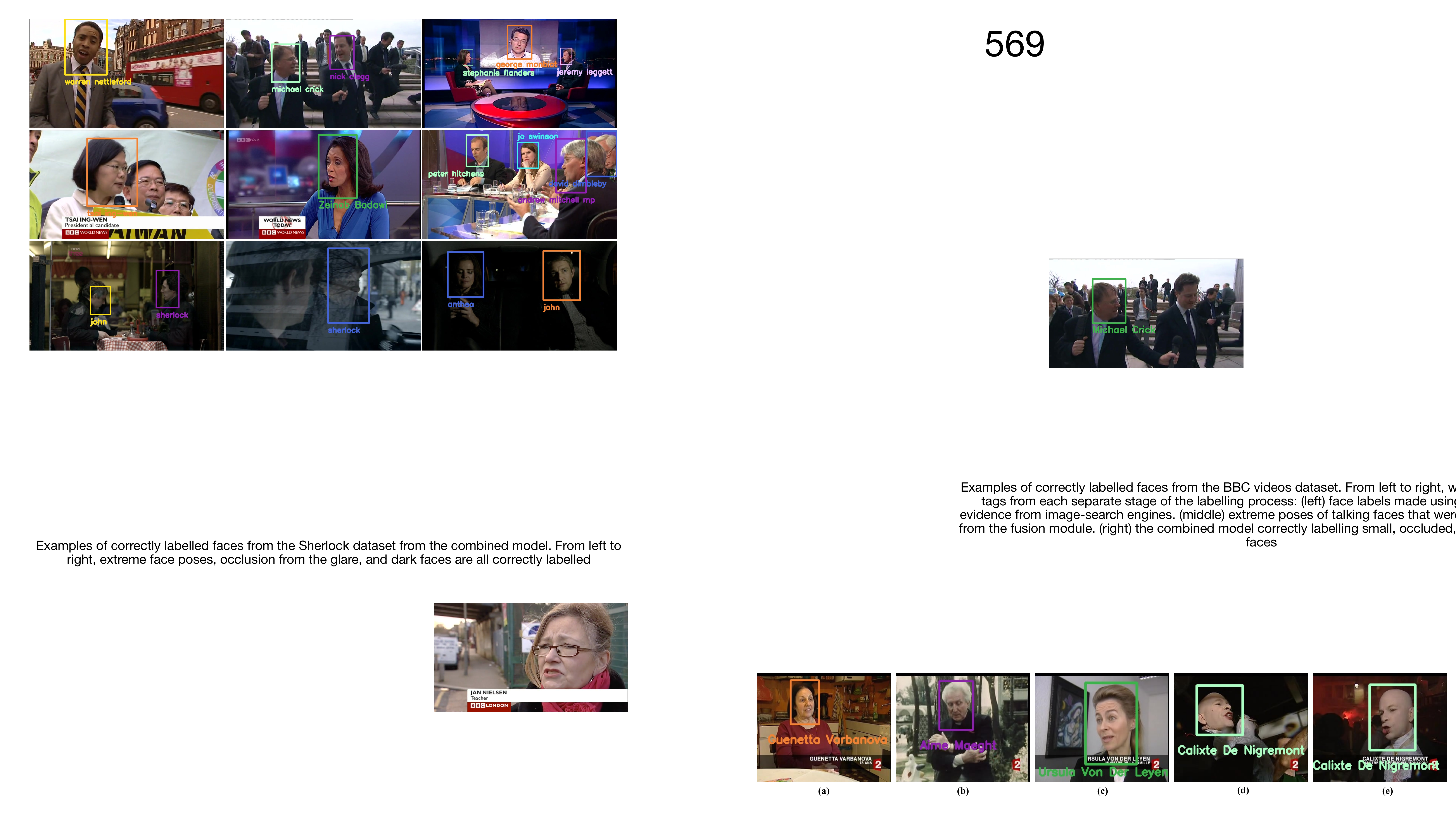}
\end{center}
\vspace{-3mm}
%\beforecaptions yeah it was bad 
\caption{Correctly labelled faces from BBC Videos (top, middle row) Sherlock (bottom row). These visually disparate datasets set challenging scenarios for person labelling, such as low resolution, lighting and extreme poses. BBC Videos: (left) Face labels obtained using \textit{corroborating evidence} from image-search engines. (middle) Talking faces labelled by the Fusion step. (right) The QE step labels small, occluded faces, and extreme poses. Sherlock: From left to right, extreme poses, occlusion from glare, and dark faces. }
%\aftercaptions
\label{fig:BBC_qual}
\vspace{-2mm}
\end{figure}

% \begin{figure}[t!]
% \begin{center}
%   \includegraphics[width=\linewidth]{figures/datasets.pdf}
% \end{center}
% % 
% %\beforecaptions yeah it was bad 
% \caption{\footnotesize{Qualitative examples of tags made by the proposed method on the 4 different datasets used in this paper. These visually disparate datasets set many challenging scenarios for person identification, such as poor resolution, lighting, and extreme face poses. From left to right: the BBC Videos dataset, MediaEval, Sherlock, IQIYI-Fusion.}}
% %\aftercaptions
% \label{qual_results}
% %
% \end{figure}
% The BBC Videos and MediaEval2015 datasets are evaluated using a retrieval protocol. The precision, recall and Average Precision are averaged across the number of different identities in the test set. The Sherlock and IQIYI-Fusion datasets are tested under the closed-set classification protocol, and so classification accuracy is recorded.} 
\subsection{Implementation Details} 
% \noindent
% \textbf{Implementation Details:} 
\label{sec:vid_pre_processing}

For automatic preparation of each dataset, we compute face-tracks (from face-detections~\cite{Chung16a}) and speech-turns (linked to faces using Active-Speaker Detection (ASD)~\cite{Chung16a} for each test-video. Additionally, we extract approximate transcripts using Automatic Speech Recognition~\cite{Afouras20} (ASR) and any scene-detected-text is found using Optical Character Recognition (OCR) techniques~\cite{deng2018pixellink,liu2018synthetically}. Full details on all video pre-processing method are given in the suppl.\ material. 

\noindent \textbf{Face and speech embeddings.} For face-tracks and speech-turns we use pre-trained embeddings to perform face~\cite{Cao18} or speaker~\cite{chung2020in} verification, respectively. For face-tracks, embeddings from each face detection are average pooled and L2-normalised into a single embedding. For speech turns, temporal average pooling of the features along the time domain produces a single utterance-level embedding.

\section{Results}
\label{sec:results}

% \new{In this section we provide an analysis into the performance of the proposed method over the 4 featured datasets. An extensive ablation study is performed on the BBC Videos dataset in Section~\ref{BBC_results}.}

% \new{As detailed in Table~\ref{tab:dataset_table}, we test different stages of the proposed method on different datasets. In this section we first give the results of the combined method on the BBC Videos and MediaEval2015 datasets, along with an extensive ablation study on the BBC Videos dataset. We then present results when elements of stage 2 are tested on the Sherlock and IQIYI-Fusion dataset.}

In this section, we first investigate the automatic method of
classifying whether a name is \textit{famous} or not, and then
evaluate either both stages, or just Stage 2, on the three datasets.

\subsection{Determining if people are \textit{famous} or not} \label{famous_non_famous_results}
\begin{figure}[t!]
\begin{center}
   \includegraphics[width=\linewidth]{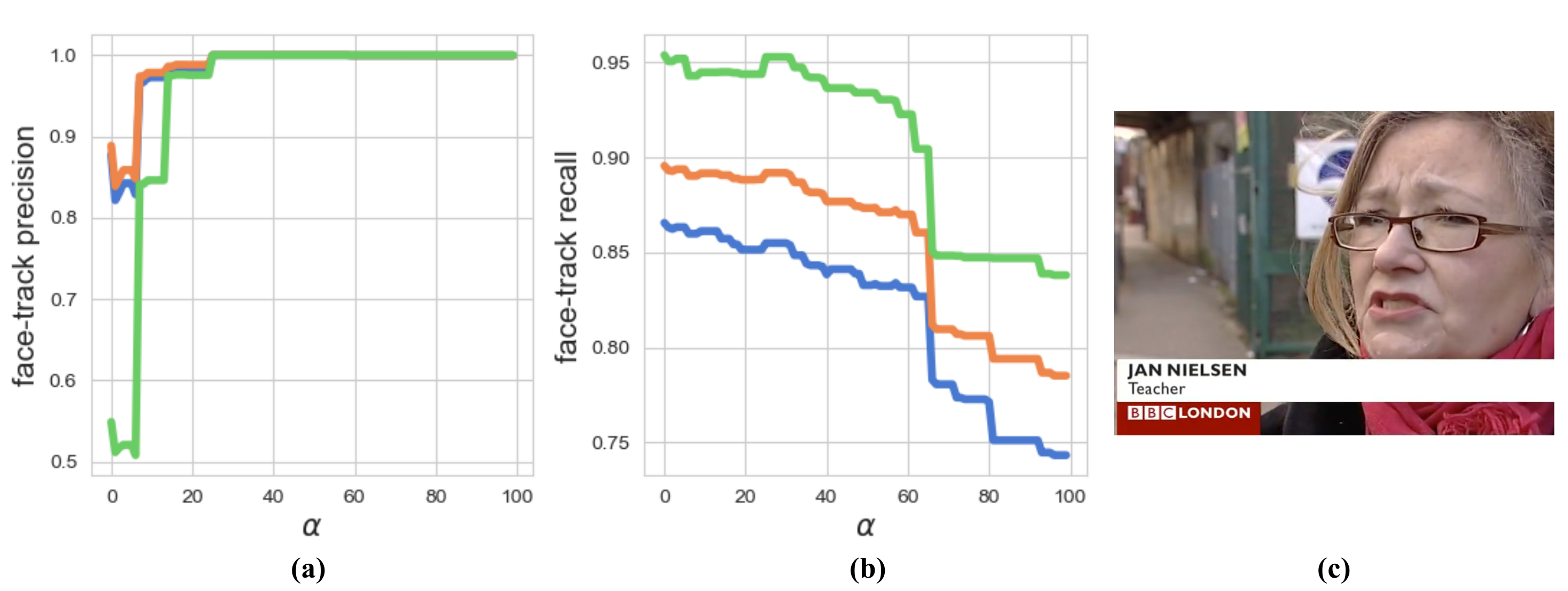}
\end{center}

\vspace{-5mm}
%\beforecaptions yeah it was bad 
\caption{ Analysis of the \textit{famous} threshold
$\alpha$ on the BBC Videos dataset.  (a) the precision of the face-track tags as
$\alpha$ increases. (b) the face-track recall as $\alpha$
increases. The colors are: blue for Stage 1,
orange for Stage 2 Fusion only, and green for Stage 2 Fusion +
QE. (c) A missed tag from the BBC Videos dataset (see Section~\ref{BBC_results}). This figure is optimally viewed in colour.}
%\aftercaptions
\label{fig:F_NF_graphs}
\vspace{-3mm}
\end{figure}

Here we investigate the choice of the \textit{famous} classification
parameter $\alpha$ on candidate names for the BBC Videos
dataset. A high $\alpha$ ($>$ 25) means that
it is likely that the faces in the largest cluster of downloaded images correspond to the searched-for person. This results in very confident face-identity models, and subsequent perfect face-track precision levels (Figure~\ref{fig:F_NF_graphs}a) when these models make correct tags.
However, a high $\alpha$ also means that the \textit{famous}
classification is limited to very well-known people. This results in low face-track recall (Figure~\ref{fig:F_NF_graphs}b), as many of the candidate names are then not
classified as \textit{famous} and so are not tagged. A low $\alpha$
($<$ 25) leads to many names being classified as \textit{famous}. This
leads to a high face-track recall ( $>$ 0.85 for Stage 1, and $>$ 0.93
for Stage 2) because face-identity models are built for many people in the videos. However,
Figure~\ref{search-engine-cluster} shows how at low $\alpha$, the largest cluster does not always depict the searched-for
person, so face-identity models
become polluted with false positives. This leads to poor face-track precision as incorrect tags are made with bad face-identity models. Stage 2 QE (blue in Figure~\ref{fig:F_NF_graphs}a/b) then creates new face models from incorrect tags, worsening the problem (precision $<$ 0.6 for $\alpha <$ 10). 

The chosen value of $\alpha=30$ reflects the balance between
achieving high face-track recall, whilst ensuring high precision. This value selects 1,967
\textit{famous} names from a total of 2,906 sourced from IMDB. 
The
IMDB names lists are not curated cast lists, and so can contain
hundreds of names irrelevant to this task. 
For written and spoken names, 
129 are classified as \textit{famous} and  170 as
\textit{less-famous}. $\alpha$ is not influenced by the video being labelled and so remains constant.

% \vspace{-6mm}

\begin{table}[t]
\caption{BBC Videos dataset results. }\label{BBC_results_table}
\centering
\begin{tabular}{|l|c|c|c|}
\hline
\multicolumn{4}{|c|}{\textbf{BBC Videos}}                                                    \\ \hline
                                  & \multicolumn{2}{c|}{Face-Tracks} &              \\ \cline{1-3}
\multicolumn{1}{|c|}{Method}      & Precision        & Recall        & Class Recall \\ \hline
Stage 1                           & 1.0              & 0.860         & 0.91         \\
Stage 1 + (Stage 2 Fusion only)      & 1.0              & 0.894         & 0.91         \\
Stage 1 + (Stage 2 QE only)          & 1.0              & 0.934         & 0.91         \\
Stage 1 + (Stage 2 Fusion + QE) & \textbf{1.0}              & \textbf{0.953}         & \textbf{0.91}         \\ \hline
\end{tabular}

\end{table}

% \begin{minipage}{\textwidth}
%   \begin{minipage}[b]{0.6\textwidth}
%   \scriptsize
%     \centering
%     \begin{tabular}{|l|c|c|c|}
% \hline
% \multicolumn{4}{|c|}{BBC Videos}                                                    \\ \hline
%                                   & \multicolumn{2}{c|}{Face-Tracks} &              \\ \cline{1-3}
% \multicolumn{1}{|c|}{Method}      & Precision        & Recall        & Class Recall \\ \hline
% Stage 1                           & 1.0              & 0.860         & 0.91         \\
% Stage 1 + (Stage 2 Fusion only)      & 1.0              & 0.894         & 0.91         \\
% Stage 1 + (Stage 2 QE only)          & 1.0              & 0.934         & 0.91         \\
% Stage 1 + (Stage 2 Fusion + QE) & \textbf{1.0}              & \textbf{0.953}         & \textbf{0.91}         \\ \hline
% \end{tabular}
%       \captionof{table}{\footnotesize{BBC Videos dataset results. }\label{BBC_results_table}}
    
%   \end{minipage}
  
%   \hspace{2mm}
  
%   \begin{minipage}[b]{0.26\textwidth}
  
%     \centering
%     \includegraphics[width=\textwidth]{figures/Jan.pdf}
%     \captionof{figure}{\footnotesize{A table besidee}}
%     \end{minipage}
%   \end{minipage}

% 
\subsection{BBC Videos} \label{BBC_results}
The BBC Videos dataset is used to test the full automated pipeline (Stages 1 and 2). 
% We first report the precision and recall across the annotated face-tracks. 
This dataset was annotated exclusively for this research. This means that we cannot compare to prior work. Instead we use the dataset to perform a quantitative analysis of the different stages of the method. Results are shown in Table~\ref{BBC_results_table}. For each face-track either the correct label is assigned, the incorrect label is assigned, or we refuse to predict a label as no model has sufficient confidence.

\noindent 
\textbf{Stage 1:} The intended design choice is to only present correctly labelled face-tracks, so the classifier threshold (for tagging faces in the test videos) is chosen on the validation set such that we make no labelling mistakes. This results in a precision of 1.0 across all episodes, meaning that no people were incorrectly tagged, whilst achieving a high face-track recall of 0.86 . A class recall of 0.91 indicates that 61 of the 66 people in the dataset were correctly tagged at least once (the missed classes are those written or spoken names for whom no
\textit{corroborating evidence} could be found on search engines). These results are impressive given that the only information provided was the programme name. Most image-search engine images are frontal faces (see Figure~\ref{search-engine-cluster}), and this is reflected in the face tags made by this stage (see Figure~\ref{fig:BBC_qual}, left column, top/middle row). As no further evidence of new, unlabelled identities is found after Stage 1, the Class Recall does not increase further. Further details are given in the suppl.\ material.

\noindent 
\textbf{Stage 2:} The fusion method improves face-track recall by 3.4\%.
Here, the voice modality is harnessed to confidently tag extreme face poses that Stage~1 could not (see Figure~\ref{fig:BBC_qual}, central column, top/middle row). 
The QE step (Stage 1 + (Stage 2 QE only)) leads to a comparatively larger 6.4\% improvement over Stage 1. QE here is able to bridge the domain gap from image-search engine images to the test video, and increase the number of correct tags. For the combined experiment (Stage 1 + (Stage 2 Fusion + QE), QE builds new identity models from the tags made by both Stage~1 and the fusion step. When combined, these offer a rich variety of poses, that are representative of the within-class variations. This therefore leads to the largest improvement to 0.953 face-track recall, where faces in a range of poses are tagged throughout the videos (see Figure~\ref{fig:BBC_qual}, right column, top/middle row).

\noindent 
Figure~\ref{fig:F_NF_graphs}c shows an example of a missed tag. This person was not labelled even though their name is displayed, as no \textit{corroborating evidence} could be found on search engines of their appearance. This problem is non-trivial, as a displayed name does not always correspond to the displayed person. Human annotators used the fact that this person is introduced in a prior scene where they were not present. For an automated process, this tag requires complex, longer term reasoning capabilities. This opens possibilities for future work. 

\subsection{MediaEval}
Our results on the MediaEval dataset are shown in Table~\ref{media_eval_res}a.
% In line with the original challenge protocol, the mean Average Precision across all featured identities is returned for this unsupervised task. 
We experiment with both the original MediaEval 2015 challenge rules~\cite{poignant2017multimodal} (no external biometric models may be used, so no image-search engines), and with the combined Stages 1 and 2 of the proposed method. Challenge participants use the strong prior that a written name is very likely to belong to the co-occurring speaking-and-visible face in the scene. Impressively our MediaEval 2015 rules method gains a 10\% improvement on the original baseline method~\cite{mediaevalgithub}, and a significant 2.2\% improvement on the state-of-the-art~\cite{poignant2014unsupervised}, while using just pre-trained out-of-the-box features. Using our full method (Stages 1 and 2),  results in an impressive further 3.88\% improvement in mAP. This gain is seen because our method is able to provide labels regardless of the proximity of a face to their corresponding written name, and is able to use stronger biometric models through the assistance of image-search engines. Our MediaEval 2015 rules method, as well as the previous baselines, fails to identify very well known people if their name is never spoken or found written. Without any manual supervision, our proposed method is able to correctly identify these people. 

\begin{table}[t!]

\caption{(a) Results on the MediaEval dataset.  (b) Results on the Sherlock dataset - values are the per-character classification accuracy for the main characters in each of the three episodes. Our method improves over the state-of-the-art for both datasets. Key: ME: MediaEval. } \label{media_eval_res}
\footnotesize
\parbox{.4\linewidth}{
\centering
\begin{tabular}{|l|c|}
\hline
\multicolumn{2}{|c|}{\textbf{MediaEval~\cite{poignant2017multimodal}} }                                  \\ \hline
\multicolumn{1}{|c|}{Method}                          & mAP (\%)       \\ \hline
Baseline~\cite{mediaevalgithub} & 74.89          \\ \hline
SOTA \cite{poignant2017multimodal}                                                   & 82.80          \\ \hline
\begin{tabular}[c]{@{}l@{}}Ours ( ME \\ 2015 rules )\end{tabular}                 & 85.22          \\ \hline
\begin{tabular}[c]{@{}l@{}}Ours (Stage 1 \\ + Stage 2)\end{tabular}            & \textbf{89.10} \\ \hline
\end{tabular}
% \begin{tabular}{l|c}
%  & MediaEval2015 \\
% \hline
% Method & mAP (\%) \\ \hline
%   Baseline~\cite{mediaevalgithub}    &  74.89   \\
%   SOA \tablefootnote{These SOA results are reported on the full version of the dataset, where as only half (the half we experimented on) has been made publicly available.} \cite{poignant2017multimodal} &  82.80 \\ \hline
%     Ours (MediaEval2015 rules)   &  85.22 \\
%     Ours (Stage 1 + Stage 2) &  \textbf{89.10}
% \end{tabular}
% 
% \subcaption{\scriptsize{ }
}
\label{sherlock_results}
\hfill
\parbox{.56\linewidth}{
\centering
\begin{tabular}{|l|c|c|c}
\hline
\multicolumn{4}{|c|}{\textbf{Sherlock~\cite{Nagrani17b}}}                                                                                                                                                  \\ \hline
\multicolumn{1}{|c|}{Method}                                                  & E01                            & E02                            & \multicolumn{1}{l|}{E03}                            \\ \hline
\begin{tabular}[c]{@{}l@{}}Nagrani \& \\ Zisserman\cite{Nagrani17b}\end{tabular} & \multicolumn{1}{c|}{0.92}      & 0.90                           & \multicolumn{1}{l|}{0.88}                           \\ \hline
Ours (face)                                                    & 0.88                           & 0.81                           & \multicolumn{1}{l|}{0.86}                           \\ \hline
%Stage 1 + (Stage 2 - Fusion)                                               & 0.89                           & 0.82                           & \multicolumn{1}{l|}{0.86}                           \\
\begin{tabular}[c]{@{}l@{}} Ours (face \\ + Stage 2)\end{tabular}                                    & \textbf{0.95} & \textbf{0.93} & \multicolumn{1}{l|}{\textbf{0.94}} \\ \hline
\end{tabular}
%
% \subcaption{\scriptsize{  }\label{Sherlockresults}}
}

\end{table} \label{media_sherlock_results}

\subsection{Sherlock}

The results for testing out Stage 2 of our method on the
Sherlock dataset are shown in Table~\ref{media_eval_res}b. The dataset
provides face images from image-search engines for each identity in the test set, so it is not necessary to run the complete Stage 1 of our method.
Instead we obtain the Stage 1 labels as follows: face embeddings are extracted for each of the provided
images, and average pooled for each identity to give a face model; 
then the cosine similarity between each of the identity models and the test video
face-tracks is computed; finally, each track is labelled with the identity and score 
of the model  that has the maximum cosine similarity.
This gives a preliminary set of face labels (`Face' result in Table~\ref{media_eval_res}b), from which Stage 2
can now boost tag numbers (`Face + Stage 2' result in Table~\ref{media_eval_res}b). Our  method
surpasses the previous state-of-the-art~\cite{Nagrani17b} considerably
by a margin of 3-6\% on all three episodes.  This is particularly
impressive, as the original work uses extra parameters for a SVM
multi-way classifier, whereas our work simply uses a nearest neighbour
classifier from aggregated identity features, which has no extra
parameters and requires no extra training. The improvements are due to superior face embeddings, and also to the evidence fusion,
which is able  to classify characters where the original work, which
treated modalities independently, could not.
Figure~\ref{fig:BBC_qual} show examples of tagged faces on this challenging dataset.

\section{Conclusions}
\label{sec:conclusions}

In this paper, we propose a novel method for the automated labelling
of people in videos through the use of \textit{corroborating evidence}, both from image-search engines, and from different information
modalities. The method performs impressively over a set of visually disparate domains both when starting from just the programme name, and also when testing out certain stages, setting new state-of-the-art results in the process. This method therefore provides a robust and reliable technique for labelling large video archives.

\section*{Acknowledgment}
\noindent
This work is supported by a EPSRC DTA Studentship, and
the EPSRC programme grant Seebibyte EP/M013774/1. We are grateful to Arhsa Nagrani, Shaya Ghadimi, and Maya Gulieva for proof reading, and the reviewers for their helpful feedback.

\bibliographystyle{IEEEtranS}
\bibliography{shortstrings,IEEEabrv,biblio_rectifier}

\end{document}